\documentclass[10pt,twocolumn,letterpaper]{article}

\usepackage{wacv}
\usepackage{times}
\usepackage{epsfig}
\usepackage{graphicx}
\usepackage{amsmath}
\usepackage{amssymb}
\usepackage{multirow}
\usepackage{float}
\usepackage{afterpage}
\usepackage{url}

\wacvfinalcopy

\begin{document}

\title{Large Scale Open-Set Deep Logo Detection}

\author{Muhammet Bastan\\
        Amazon\\
        Palo Alto, CA, USA\\
        {\tt\small mbastan@amazon.com}
\and
    Hao-Yu Wu\\    
    Pinterest\\
    Palo Alto, CA, USA\\
    {\tt\small rexwu@pinterest.com}
\and
	Tian Cao\\
	Houzz\\
	Palo Alto, CA, USA\\
	{\tt\small tian@houzz.com}    
\and
     Bhargava Kota\\
     SUNY University at Buffalo\\
     Buffalo, NY, USA\\
     {\tt\small buralako@buffalo.edu}     
\and
    Mehmet Tek\\
	Google\\
	Mountain View, CA, USA\\
	{\tt\small mtek@google.com}
}

\maketitle
\ifwacvfinal\thispagestyle{empty}\fi

\begin{abstract}
   We present an open-set logo detection (OSLD) system, which can detect (localize and recognize) any number of unseen logo classes without re-training; it only requires a small set of canonical logo images for each logo class.
   We achieve this using a two-stage approach: (1) Generic logo detection to detect candidate logo regions in an image. (2) Logo matching for matching the detected logo regions to a set of canonical logo images to recognize them.
   
   We constructed an open-set logo detection dataset with 12.1k logo classes and released it for research purposes.We demonstrate the effectiveness of OSLD on our dataset and on the standard Flickr-32 logo dataset, outperforming the state-of-the-art open-set and closed-set logo detection methods by a large margin. OSLD is scalable to millions of logo classes.
\end{abstract}

\section{Introduction}
\begin{figure}
	\centering	
	\includegraphics[width=0.5\textwidth]{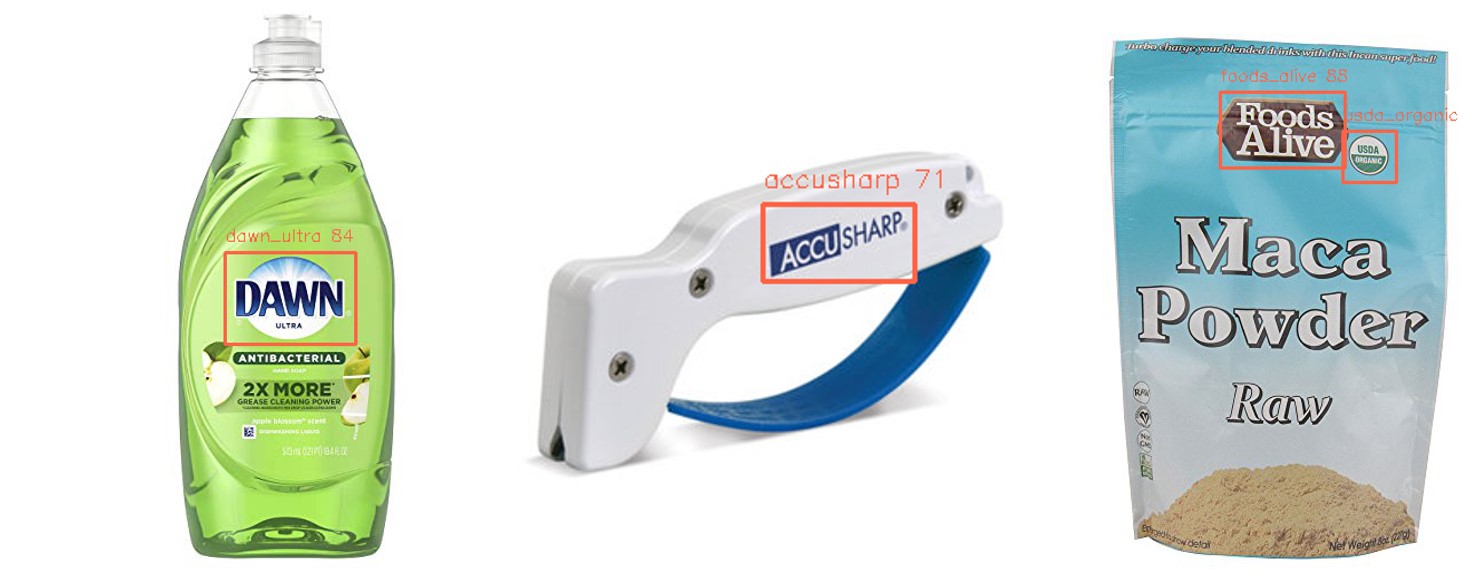}	
	\caption{\footnotesize Our system can detect logos (localize with a bounding box and identify its brand) in images even when the logo class is not present in the training set (logo detection). It only needs a set of canonical logo images for each logo class (Figure~\ref{fig:logo}).}
	\label{fig:intro}
\end{figure}

Logo detection is the task of localizing and identifying logos in images (Figure~\ref{fig:intro}), with practical applications such as brand protection, brand-aware product search and recommendation. It is a difficult task, even for humans, since it is usually hard to tell whether a graphics pattern or piece of text belongs to a logo or not without prior knowledge of the original logo. Well-known logos are usually easy to spot and identify using prior knowledge, while unknown logos may be hard to recognize.

Traditionally, logo detection is treated as a \textit{closed-set} object detection problem, in which the system is trained for a predefined set of logo classes and can only recognize those classes at test time. It has been shown to work well for a small number of logo classes~\cite{deep-logo-15,dl-logo-nc17}, e.g., 32 logo classes in the Flickr-32 logo dataset~\cite{flickr-32}, given sufficient number of training samples for each class. However, \textit{closed-set} logo recognition has the following major shortcomings that makes it unsuitable for large-scale real world logo detection.

\begin{itemize}
	\item It can not recognize new logo classes without re-training, which is expensive.
	\item It is not scalable to a realistic number of logo classes (METU logo dataset~\cite{metu-ds15} has ~410K logo classes).
	\item It is very expensive to gather and annotate sufficient number of logo images to properly train the detection/recognition systems.
	\item Object detectors with classification heads are not scalable to large number of classes.
\end{itemize}

\begin{figure}
	\centering
	\includegraphics[width=0.5\textwidth]{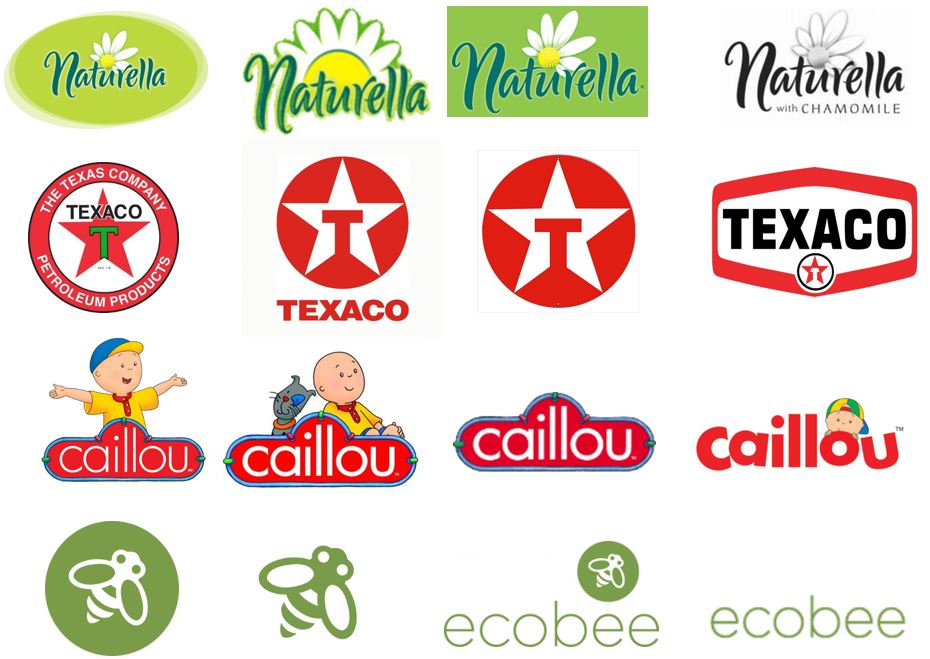}	
	\caption{\footnotesize Canonical logo images. There might be significant differences between the versions of the same brand logo.}
	\label{fig:logo}
\end{figure}

\textit{Open-set} logo detection~\cite{open-set18,proxy-wacv19} addresses these shortcomings. It can detect and recognize new logo classes without re-training. It is scalable to large number of logo classes. There is no need to annotate large  number of images for each and every new logo class, it is sufficient to provide only the canonical logo images (Figure~\ref{fig:logo}) for the logo classes to be recognized. Therefore, open-set logo detection is much better suited to real-world logo detection scenarios, where there are a large number of brand logos which might be changing over time and new logo classes might be arriving continuously.

Motivated by this, we present an \textit{open-set} logo detection system, inspired by the recent success of similar systems for face recognition~\cite{Wang2018} and person re-identification~\cite{Wu2019} tasks. Logo detection is similar in that we can formulate it as a two-stage problem: generic logo detection, followed by logo matching, which is feasible as there are typically a small number of canonical logo images for each brand logo (Figure~\ref{fig:logo}).

This work has the following contributions:
\begin{itemize}
	\item An end-to-end open-set logo detection system (OSLD) that outperformed previous open-set and closed-set methods by a large margin.
	\item A new logo detection dataset with 12.1k logo classes (Section~\ref{sec:dataset}), released for research purposes. 
\end{itemize}

\section{Related Work}
\label{sec:related}
Logo detection is a special case of object detection, which comprises object localization and recognition. Traditionally, logo detection is treated as a \emph{closed-set} object detection problem, in which the system is trained on a small number of pre-defined logo classes. Early systems relied on hand-engineered local keypoint features~\cite{Joly2009,Yannis2011,flickr-32,Romberg2013}. Recently, deep learning based methods have been dominant in closed-set logo detection~\cite{Bianco2015,Eggert2015,Bao2016,Oliveira2016,Su2016,Su2017,Bianco2017}. They trained Faster R-CNN~\cite{faster-rcnn15} object detector, or used logo region proposals followed by classification with CNN features. These methods require a large amount of memory for large number of output classes and cannot detect new logo classes without re-training. It is also costly to gather and annotate sufficient amount of training data. In~\cite{Su2017}, the authors present a partial solution to the dataset construction problem. They start with a small amount of training data to train a model and progressively expand the dataset by running the model on web images and retaining the high confidence detections (semi-supervised learning). This way, they built the WebLogo-2M dataset with 1.9M images, but with only 194 logo classes.

\emph{Open-set} logo detection takes a two-stage approach: generic logo detection followed by logo matching using learned CNN embeddings. This is similar to the recent deep learning based face recognition~\cite{Wang2018} and person re-identification~\cite{Wu2019} approaches, which achieved impressive performance. In~\cite{open-set18}, Faster R-CNN is used to first localize candidate logo regions. Then, another CNN, pre-trained on ImageNet and fine-tuned on the logo dataset is used to extract features from the candidate logo regions and match to the database. In~\cite{proxy-wacv19}, the authors employed the so-called Proxy-NCA metric learning approach to learn better embeddings to match candidate logo regions to canonical logo images. Trained on a large (295K product images with 2K logo classes, downloaded from the web), they also demonstrated $+10$ points improvement in mAP on the Flickr-32 dataset over the previous works, without re-training or fine tuning. Our framework (Figure~\ref{fig:framework}) is similar, but with better metric learning and training strategies, outperforming the previous methods by a large margin.

\section{Method}
\begin{figure*}
	\centering
	\includegraphics[width=1.0\textwidth]{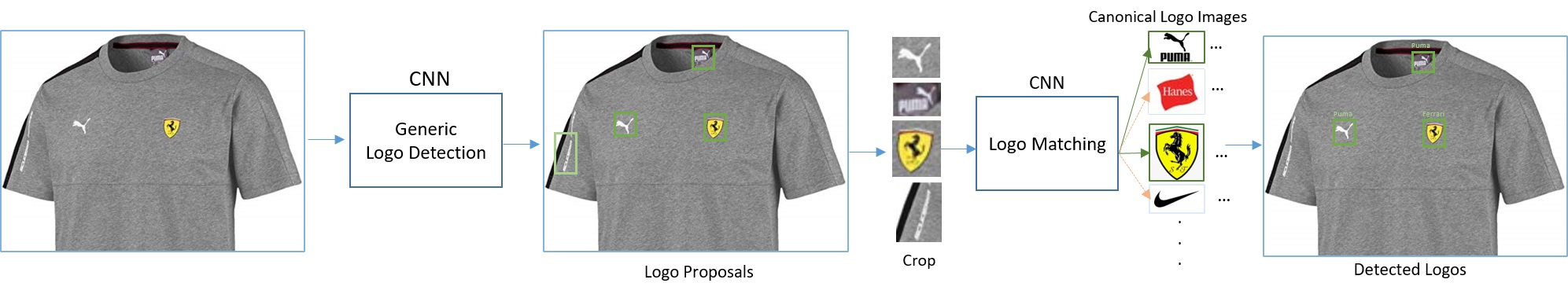}
	\caption{\footnotesize Our  logo detection framework. Generic logo detector outputs candidate logo regions, which are cropped from the image and then matched to a set of canonical logo images using low dimensional CNN embeddings learned for matching logo images.}
	\label{fig:framework}
\end{figure*}

Inspired by the high performance of two-stage deep metric learning based approaches, as in face recognition and person re-identification, we take a two-stage approach to  logo detection, as shown in Figure~\ref{fig:framework}. The first stage, logo detection (Section~\ref{sec:logo-det}), localizes candidate logo regions using a generic logo detector, which should have high recall, but may have low precision. The second stage, logo matching (Section~\ref{sec:logo-match}), matches the candidate logo regions to the set of all canonical logo images for all logo classes to be recognized. The matching module is trained to assign high score to the correct matches and low score to the false matches. The detected logo regions are labeled with the best matching logo class.

Whenever a new logo class needs to be recognized, it is sufficient to add its canonical logo images (CNN representations) to the logo database and no further data collection, annotation and training are required. This flexibility is very useful in real-world logo detection applications.

\subsection{Generic Logo Detection}
\label{sec:logo-det}
Generic logo detection localizes candidate logo regions with bounding boxes, without identifying their class. Hence, a binary logo detector with two outputs (logo, background) is sufficient. The detector should have high recall. Due to the ambiguity of the logo detection problem, we expect to detect false logo candidates, typically text, graphics or graphics+text regions, which are filtered out in the second stage when they can not be matched to any canonical logo image in the database.

We used the RFBNet~\cite{rfbnet} object detector due to its high speed and accuracy, with $500 \times 500$ input resolution (most of the images in our dataset has this or lower resolution) and VGG16 base network. It is a single stage object detector based on SSD~\cite{ssd}. We trained the RFBNet using a logo dataset labeled with bounding boxes. The class labels are not needed, as we only need a binary (generic) logo detector. This makes the annotation much cheaper, compared to closed-set approaches which also need the class labels for each bounding box.

We tuned the anchor box sizes and aspect ratios of the RFBNet based on the bounding box statistics of the training set. We used a pre-trained VGG16 network, first frozen the base convolutional layers and trained only the additional layers with Adam optimizer with a learning rate of $10^{-4}$, then trained the whole network with an initial learning rate of $10^{-5}$, reduced at $10^{th}$ and $30^{th}$ epochs by half, for 50 epochs. We saved the best model with the best validation accuracy measured by mean average precision (mAP).

\subsection{Logo Matching}
\label{sec:logo-match}
The goal of logo matching is to match the candidate logo regions from the generic logo detection to canonical logo images in the database, to identify the class of each logo region or discard it if it does not match to any logo image with high confidence. In Figure~\ref{fig:framework}, the generic logo detector localized four candidate logo regions, three of which are true logo regions and one is false. The logo matching module takes the cropped candidate logo regions and tries to match each one to all the canonical logo images in the database, using their $L_2$-normalized CNN embeddings and Euclidean or cosine distance. The best matching canonical logo class is assigned as the label (`Puma', `Ferrari' in the figure) if the matching score is above a threshold. The non-logo regions are discarded as they do not match to any canonical logo image with high score. The CNN embedding network for logo matching is trained with deep metric learning (DML) losses, as described in the next section.

\section{Deep Metric Learning}
\label{sec:dml}
Metric learning for images aims to learn a low dimensional image embedding that maps similar images closer in the embedding space and dissimilar ones farther apart. Deep metric learning (DML) using CNN embeddings has achieved high performance on image retrieval~\cite{bier-pami18,abe2018}, face recognition ~\cite{Wang2018}, person re-identification~\cite{Wu2019}, to name a few.

There are mainly two types of loss functions to train the DML networks: pairwise and proxy (proxy-NCA, proxy anchor) loss functions. Pairwise losses (e.g., contrastive, triplet) are computed between positive and negative pairs. Proxy losses (e.g., proxy-NCA, proxy-anchor) learn a set of embeddings, called proxies, to represent the class centroids; the loss for each training sample is computed with respect to the proxies.

Triplet and contrastive losses are commonly used pairwise DML loss functions, although their gradients are not continuous. Binomial deviance loss~\cite{bier-pami18,sgml19} is similar to contrastive loss, but with continuous gradients. Here we consider the triplet and binomial deviance loss functions.

Let $(x_a, x_p, x_n)$ be a triplet of anchor, positive and negative images, and let $(f(x_a), f(x_p), f(x_n))$ be their CNN embeddings, e.g., the output of the last fully connected layer. Following~\cite{bier-pami18,sgml19}, we use the cosine similarity, $s$, between the image embeddings $f(x_i)$.

\newcommand{\norm}[1]{\left\lVert#1\right\rVert}
\begin{equation}
	s(f(x_1), f(x_2)) =  \frac{f(x_1)^{\top}f(x_2)}{\norm{f(x_1)} \cdot \norm{f(x_2)}}
\end{equation}

The cosine similarity lies in $[-1,+1]$ and enforces an upper bound on the loss value, possibly helping the optimization process.

The \textbf{triplet loss}, $L_t$, is defined as follows.

\begin{equation}
	L_t = max(0, s_{an} - s_{ap} + m)
\end{equation}

where, $m$ is the margin value, $s_{ap}$ and $s_{an}$ are cosine similarities between anchor-positive and anchor-negative image embedding pairs.

Similarly, the \textbf{binomial deviance loss} (bindev), $L_{bd}$, is defined as follows.

\begin{align}
	L_{ap} &= \log(1 + e^{- \alpha (s_{ap} - m) }) \\
	L_{an} &= \log(1 + e^{\alpha (s_{an} - m) }) \\
	L_{bd} &=  L_{ap} + L_{an}
\end{align}

where, $\alpha$ and $m$ are scaling and translation (margin) parameters (which can also be different for positive and negative pairs and should be tuned for each dataset and model). This is slightly different from the formulation in~\cite{bier-pami18}, which uses a cost parameter $C_y$ to balance the positive and negative pairs in the loss. We omit this term as we use a balanced sampling strategy, as explained in the next section.

\subsection{Simple Deep Metric Learning (SDML)}
\label{sec:sdml} 
We now present some simple tricks or best practices that we empirically found to work well in deep metric learning and enable our simple deep metric learning (\textbf{SDML}) framework to achieve high accuracy, on par or better than state-of-the-art.

\begin{itemize}
	\item \textbf{Sampling.} Sampling good pairs/triplets is known to be very important in training deep metric networks~\cite{sampling17}. Random sampling is simple, but leads to easy pairs which are not helpful in the learning process. Hard negative mining (HNM) is a commonly used solution in such cases. However, HNM is typically applied within a batch, which is sampled randomly. We recommend balanced random sampling followed by hard negative mining (BHNM). First sample $N$ anchor images, and $N$ positive and $N$ negative images (total $3N$). Then, consider all pairs of positives and negatives withing the batch and apply hard negative sampling. This strategy produces equal number of positive and negative pairs (hence no need to use a balancing factor in the loss function as in~\cite{bier-pami18}), and makes better use of the available data.
	
	\item \textbf{Preprocessing and Data Augmentation.} The input images to CNNs are typically resized to square size, e.g., $224 \times 224$, for efficient mini-batch processing. Directly resizing to square size may lead to large distortions in the image when the aspect ratio is either large or small. In such cases, padding the input image to square size should be preferred to avoid distortion, and this affects the performance significantly.
	
	For data augmentation, random resized crop and flipping are typically used. The parameters of random resized crop should be tuned carefully according to the dataset, otherwise it will adversely affect the performance, e.g., including very small scale crops which may end up being the background or cause large object scale discrepancy between training and test sets, or wide range of aspect ratios which will heavily distort the images. Training and test resolutions should be similar for optimal performance~\cite{fixres19}.
	
	\item \textbf{Optimization.} Stochastic gradient descent (SGD) and Adam are commonly used optimizers in DML. We recommend the Adam optimizer with an initial learning rate of around $10^{-4}$ and decreasing learning rate scheduler, which we found to provide fast and stable training with good results. Larger learning rates may lead to unstable training and/or inferior performance; smaller learning rates, e.g., $10^{-6}$, will need long training times and may get stuck at local minima.
	
\end{itemize}

\subsection{SDML Results} 
Now, we demonstrate the effectiveness of our SDML framework by comparing it with the state-of-the-art DML methods. We performed experiments on one of the standard deep metric learning datasets, SOP~\cite{SOP2016}, which contains e-commerce product images, similar to our logo dataset.

\begin{table*}[h]
	\centering
	\small
	\begin{tabular}{l c c c c c}
		\hline
		Method 									& Base CNN 	 & Image Size & Emb Size 	& R@1 \\
		\hline		
		A-BIER~\cite{bier-pami18} (PAMI 18)  	& GoogleNet  & 224 & 512 & 74.2  \\
		ABE~\cite{abe2018} (ECCV 18)		    & GoogleNet   & 224 & 512 & 76.3 \\ 
		XBM~\cite{xbm20} (CVPR 20)		    & GoogleNet   & 224 & 512 & 77.4 \\ 
		\hline
		SDML (bindev)		                    & GoogleNet   & 224 & 512 & \textbf{77.7}  \\
		\hline
		Margin~\cite{sampling17} (ICCV 17)      & ResNet50   & 256 & 128 & 72.7  \\
		Smooth-AP~\cite{sap20} (ECCV 20)      & ResNet50    & 224 & 512 & 80.1 \\
		Proxy-Anchor~\cite{pa20} (CVPR 20)      & ResNet50    & 224 & 512 & 80.5 \\
		XBM~\cite{xbm20} (CVPR 20)      & ResNet50    & 224 & 128 & 80.6 \\
		Proxy-NCA++~\cite{pnca20} (ECCV 20)      & ResNet50    & 224 & 512 & 80.7 \\
		\hline
		SDML (bindev)		             & ResNet50  & 224 & 128 & \textbf{80.9} \\
		SDML (bindev)		             & ResNet50  & 224 & 512 & \textbf{81.3} \\
		SDML (bindev)		             & DenseNet169  & 224 & 128 & \textbf{81.5} \\
		SDML (bindev)		             & DenseNet169  & 224 & 512 & \textbf{83.1} \\
		SDML (bindev)		             & DenseNet169  & 224 & 1024-512 & \textbf{84.0} \\
		\hline	
	\end{tabular}
	\caption{\footnotesize Comparison of our SDML with state-of-the-art methods on SOP dataset. The methods are grouped by base CNN architecture and embedding size. We re-trained proxy anchor loss with ResNet 50 using the SDML settings.}
	\label{table:sdml_sop}			
\end{table*}

We first briefly review the latest state-of-the-art methods on DML, and then compare our SDML with them. In~\cite{sampling17}, a distance weighted sampling strategy was proposed to select more informative training examples and was shown to improve retrieval. Opitz et al.~\cite{bier-pami18} introduced the binomial deviance loss to deep metric learning, as an alternative to triplet loss. They also proposed a DML network, BIER/A-BIER, with embedding ensemble, online gradient boosting and additional diversity loss functions. Later, Kim et al.~\cite{abe2018} improved the ensemble model of BIER with attention, attention based ensemble (ABE), to attend to different parts of the image and a divergence loss to encourage diversity among the learners. 

More recently, cross-batch memory (XBM)~\cite{xbm20} uses additional memory for the already processed images to sample more informative pairs. Smooth-AP loss~\cite{sap20} optimizes the average precision metric explicitly. Proxy-Anchor loss~\cite{pa20} uses the proxies as anchors, and ProxyNCA++~\cite{pnca20} is a set of tricks to train with the ProxyNCA~\cite{pnca17} loss.

\noindent
\textbf{SDML Setting.} We grouped the existing DML methods based on their base CNNs and embedding size and used the same settings for a fair comparison. We use ImageNet pre-trained base CNNs. We first fine tuned the embedding layer for $15$ epochs with a learning rate of $10^{-4}$, and then fine tuned the whole network for a maximum of 30 epochs, with an initial learning rate of $10^{-4}$, halved when validation accuracy stopped improving. We used Adam optimizer; $\alpha=3.0$ and $m=0.3$ in the loss functions. With the Adam optimizer, initial learning rate in the order of $10^{-4}$ works well across different CNNs and datasets. Following common practice, we evaluated at every epoch and reported the best accuracy (there is no validation set in any of the four standard DML datasets), which we observed to be repeatable over multiple runs.

We used the balanced random sampling with hard negative mining (BHNM) as described above with $N=60$. We padded the input images with zeros to square size and used random horizontal flip and random resized crop with scale in [0.8, 1.0], aspect ratio in [0.9, 1.1] as data augmentation. At test time, we padded to square size with zeros and resized to $224 \times 224$.

\noindent
\textbf{Comparison.} Table~\ref{table:sdml_sop} compares our SDML with the state-of-the-art DML methods reviewed above, using the commonly used Recall@K performance measure (percentage of queries with at least one relevant result in top K). The results are grouped and sorted by base CNN and embedding size.

With GoogleNet, SDML is significantly better than A-BIER~\cite{bier-pami18} (+3.4) and ABE~\cite{abe2018} (+1.3), and on par with XBM which uses additional memory.

With ResNet50, SDML is better than Smooth-AP~\cite{sap20} (+1.2), Proxy-Anchor~\cite{pa20} (+0.8) and ProxyNCA++~\cite{pnca20} (+0.6), and on par with XBM~\cite{xbm20}, which uses additional memory. SDML obtains even better results with DenseNet169, even though DenseNet169 has much fewer parameters than ResNet50 (14M vs 25M). We gain another +0.9 points by adding a fully connected layer of size 1024 before the embedding layer and with larger batch size ($N=220$). In summary, the recent loss functions perform similarly under the same settings, and better models provide more gains, i.e., it is possible to obtain much better results with the latest state-of-the-art CNN/transformer models.

Next, we return to our original problem of open-set logo detection and employ our findings in SDML to boost logo matching accuracy.

\section{Datasets}
\label{sec:dataset}

We need a large scale dataset with a large number of logo classes to evaluate our  logo detection framework. There is no such public dataset, therefore, we  built our own dataset, OSLD, which we released for research purposes\footnote{Available at \url{https://github.com/mubastan/osld}}.

\begin{figure}
	\centering
	OSLD dataset\\
	\includegraphics[width=0.5\textwidth]{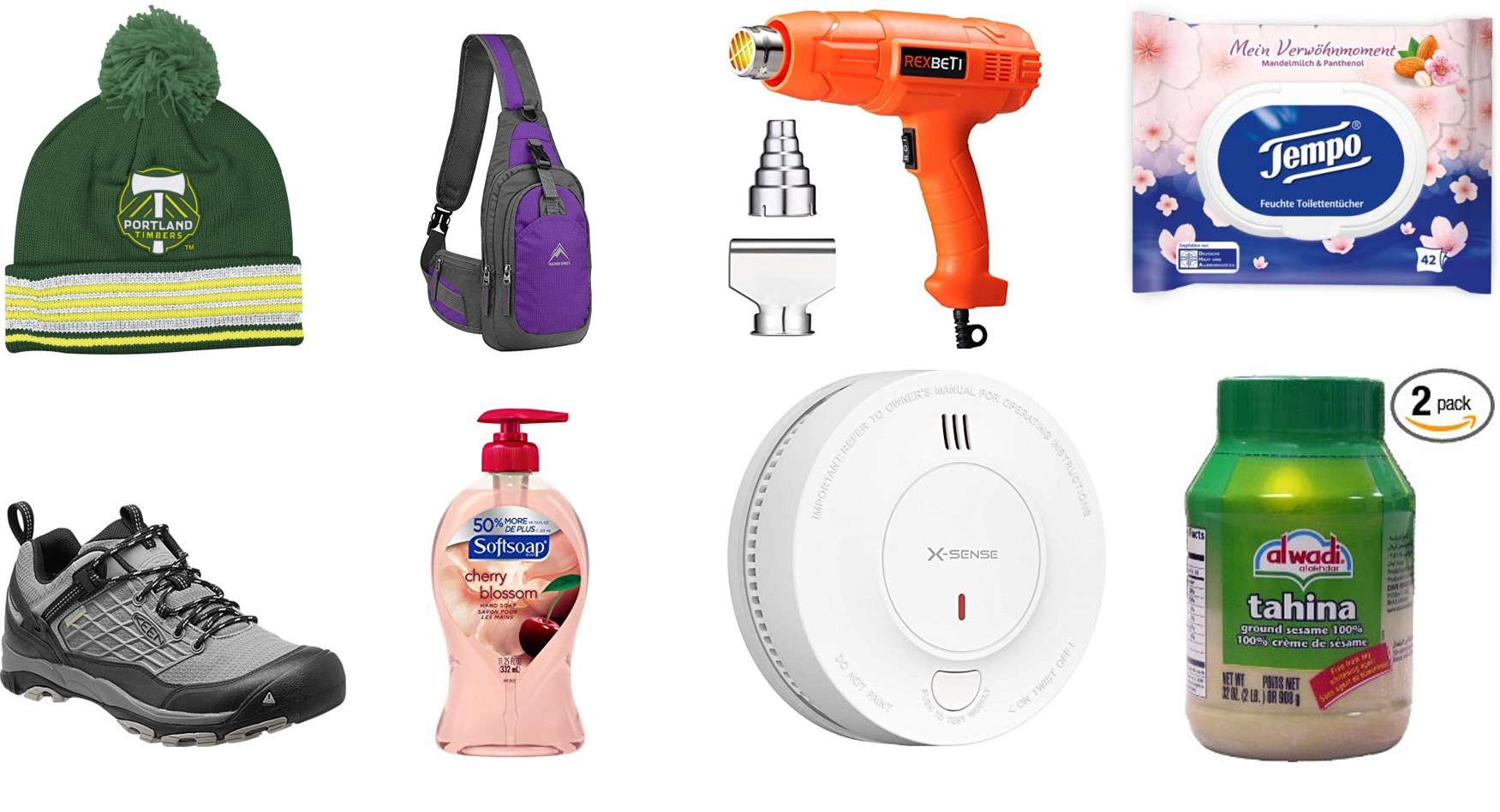}
	
	\vspace{1em}
	Flickr-32 dataset\\
	\includegraphics[width=0.5\textwidth]{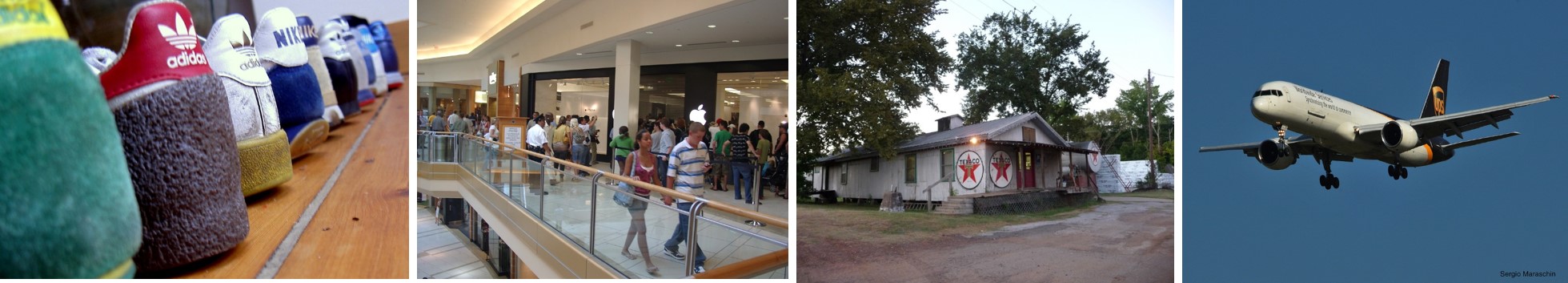}	
	
	\caption{\footnotesize Sample images from the OSLD and Flickr-32 logo datasets.}
	\label{fig:samples}
\end{figure}

\noindent
\textbf{OSLD (Open Set Logo Detection) Dataset}. OSLD dataset consists of eCommerce product images, as shown in Figure~\ref{fig:samples}. The dataset has 20K labeled images, 1M unlabeled image URLs, and 12.1K logo classes with 20.8K canonical logo images. We obtained two types of annotations to train our detection and matching networks: 
\begin{itemize}
	\item Logo bounding boxes (bbox), without class labels, for  20K product images with 12.1K logo classes. We used these annotations to train and evaluate the generic logo detection. Note that it would not be possible to train a closed-set logo detection system on this dataset, as there are only a few instances of each logo class.
	\item The matching canonical logo image for each of the bounding boxes. We used these bounding box to canonical logo image pair annotations to train the matching CNN.
\end{itemize}

We split the dataset into training, validation and test sets, in a strictly open-set setting, i.e., training, validation and test set logo classes are totally different and there is no logo class overlap between the splits. The training, validation and test sets have 15.7K, 1.8K and 2.4K images with 4.1K, 500 and 500 logo classes respectively. The splits contain 300, 200 and 200 images without any logos.

The 20K labeled product images are for supervised training of logo detection and matching models, whereas the 1M unlabeled product images are provided for self/semi-supervised training. In this work, we did not train on the 1M unlabeled images, but still released them to promote research in self/semi-supervised logo detection.

\noindent
\textbf{Flickr-32 Logos}. This is a commonly used logo dataset with 32 logo classes~\cite{flickr-32}. Released in 2011, prior to deep learning era, it has a tiny training set of 320 images; test set has 3960 images, 960 of which contain logos.  We used this dataset to compare to earlier logo detection methods and also measure how transferrable our logo detection system is to other datasets.

As can be seen in Figure~\ref{fig:samples}, the two datasets are quite different in terms of image content, clutter, logo size and transformations.

\section{Results}
\label{sec:results}

\subsection{Generic Logo Detection}
We trained and evaluated the generic logo detection network (RFBNet) on the OSLD detection training set (15.5K), as described in Section~\ref{sec:logo-det}.
We evaluated the performance using the standard PASCAL VOC object detection evaluation procedure with mean average precision (mAP) and recall as the performance measures: a detection is correct if it has at least 50\% intersection over union (IoU) with a ground truth bounding box. We evaluated on OSLD test set and also Flickr-32 test set; Table~\ref{table:det} shows the results. On OSLD test set, mAP is $76.3$, and recall is $96.2$.

Using the RFBNet model trained on OSLD training set, the mAP on Flickr-32 test set is $42.9$, which is much lower than that on OSLD test set, but recall ($76.9$) is still quite good. We also fine tuned the OSLD RFBNet model on the tiny Flickr-32 training set (with only 320 images), the mAP improved to $58.3$ and recall to $83.3$. Based on these results, we conclude that there is a significant difference between the two datasets, in terms of generic logo detection.

\begin{table}[h]
	\centering
	\begin{tabular}{l c c }
		\hline
		Dataset & mAP	& Recall\\
		\hline
		OSLD & 76.3  & 96.2\\
		Flickr-32 & 42.9  & 76.9\\
		Flickr-32 $^*$ & 58.3  & 83.3\\
		\hline	
	\end{tabular}
	\caption{\footnotesize Generic logo detection (localization) test set performance with RFBNet~\cite{rfbnet} using PASCAL VOC 50 \% IoU criterion, trained on OSLD training set. $^*$RFBNet fine tuned on Flickr-32 training set (320 images).}
	\label{table:det}
\end{table}

\subsection{End-to-End Logo Detection}
Here, we present the performance of the end-to-end detection: given an input image, RFBNet localizes candidate logo regions, which are matched to the canonical logo images using SDML logo matching model, as shown in Figure~\ref{fig:framework}. For each candidate logo region, we consider only the top-1 match returned by logo matching. We use standard PASCAL VOC object detection evaluation procedure with mAP measure: a detection is correct if it has at least 50\% IoU with a ground truth bounding box and its label is correct. We also report image-based mAP, to compare with the earlier methods, all of which used image-based mAP, in which a detection is correct if its label matches to one of the ground truth labels in the image; for each logo class we take the detection with the maximum score (single detection per logo class).

Note that the evaluations on both OSLD and Flickr-32 are done in a strictly open-set setting, i.e., the test set logo classes were never seen during training.

Based on the SDML experiments, we selected the DenseNet169 model with one extra FC layer of size $1024$ and embedding size $512$, and input image size $227 \times 227$.
To account for the color variations of the logos, we additionally used color jitter (hue=0.1, brightness=0.1, contrast=0.1, saturation=0.1) and random grayscale conversion as data augmentation. Moreover, to account for the false positive logo candidates during matching, we included the false positives from generic logo detection as negative pairs during triplet sampling. Other training parameters are the same.

\noindent
\textbf{Results on OSLD.}
Table~\ref{table:osld-res1} shows the end-to-end logo detection and matching performances on OSLD test set (2.5K images with 500 logo classes) for two matching methods, SDML with triplet loss and binomial deviance loss. Even though there are only 500 logo classes in the test set, we also presented the results for the entire 12K logo classes (with 20.5K images) during matching, which is more challenging than matching to only 500 logo classes. SDML with binomial deviance loss worked slightly better than triplet loss (about +2pps) and matching performance for 500 classes is much higher (about +10pps) than 12141 classes, as expected. Table~\ref{table:osld-res2} shows the results with the ground truth bounding boxes, instead of the RFBNet bounding boxes. Again, as expected, the performance is much better with ground truth bounding boxes. This means that there is a lot of room for improvement with better logo localizers (better object detectors trained for generic logo detection).

\begin{table}[h]	
	\centering
	\begin{tabular}{l c c c }
		\hline
		Method & \# classes	& Image-based	& BBox-based \\
		\hline
		SDML (triplet) & 12141 &  74.6 & 68.2  \\
		SDML (triplet) & 500 & 85.2 & 78.0  \\
		SDML (bindev) & 12141 & 77.1  & 70.4  \\
		SDML (bindev) & 500 & 86.3  & 78.9  \\				
		\hline	
	\end{tabular}
	\caption{\footnotesize End-to-end image-based and bbox-based detection performance (average precision -- AP) on the OSLD test set (2.5K images with 500 logo classes) for 12141 and 500 logo classes in the logo database. Each candidate logo region is matched to all 12141 or 500 logo classes. SDML: Simple Deep Metric Learning (Section~\ref{sec:sdml}).}
	\label{table:osld-res1}
\end{table}

\begin{table}[h]	
	\centering
	\begin{tabular}{l c c c }
		\hline
		Method	&\# classes	& Image-based	& BBox-based \\
		\hline
		SDML (triplet) & 12141 &  86.7 & 83.0  \\
		SDML (triplet) & 500 &  93.8 & 90.7  \\
		SDML (bindev) & 12141 & 89.3  & 85.4  \\
		SDML (bindev) & 500 & 95.1  & 92.1  \\				
		\hline	
	\end{tabular}
	\caption{\footnotesize Image-based and bbox-based detection performance (average precision -- AP) on the OSLD test set with the ground truth bounding boxes (instead of RFBNet detections) and for 12141 and 500 logo classes. SDML: Simple Deep Metric Learning (Section~\ref{sec:sdml}).}
	\label{table:osld-res2}
\end{table}

\begin{table*}[!htb]	
	\centering
	\small
	\begin{tabular}{|l| l c c c|}
		\hline
		& Method	&  Training Data & Image-based AP	& BBox-based AP \\
		
		\hline
		\multirow{3}{*}{Closed-Set}
		
		& Fast R-CNN~\cite{Oliveira2016} & Flickr-32 & 73.5 & -\\
		& Faster R-CNN~\cite{Su2016} & Flickr-32 + Synthetic & 81.1 & -\\
		& Faster R-CNN~\cite{Bao2016}  & Flickr-32 & 84.2 & -\\
		
		\hline
		\multirow{6}{*}{Open-Set}
		
		& Proxy-NCA (top-1)~\cite{proxy-wacv19} & PL2K & 44.4  & -\\
		& Faster R-CNN~\cite{open-set18} & Flickr-32 + LitW & 46.4 & -\\
		& Proxy-NCA (top-5)~\cite{proxy-wacv19} & PL2K & 56.6  & -\\
		
		\cline{2-5}
		& SDML (triplet) & OSLD & 90.2  & 52.7 \\ 
		& SDML (bindev)  & OSLD & \textbf{91.2}  & 52.9 \\  	
		& SDML$^*$ (triplet) & OSLD &  \textbf{91.3} & 60.6 \\
		& SDML$^*$ (bindev) & OSLD & 91.3  & 60.3 \\ 	
		\hline	
	\end{tabular}
	\caption{\footnotesize Comparison of our open-set logo detection method (OSLD/SDML) with the state-of-the-art open-set and closed-set logo detection methods on the Flicker-32 test set. All values are AP (average precision). $^*$Our RFBNet generic logo detector was fine-tuned on Flickr-32 training set.}
	\label{table:flickr-res}
\end{table*}

\noindent
\textbf{Results and Comparison on Flickr-32.} We performed experiments on Flickr-32 test set (3960 images, 32 logo classes), using the models trained on OSLD training set (SDML) and compared with the state-of-the-art open-set and closed-set logo detection methods.
There is no overlap between the logo classes of OSLD training set and Flickr-32 test set (open-set setting). For the sake of fair comparison with previous works, we used the 32 Flicker-32 logo classes as match set (in contrast to 12.1K in OSLD test).
The results and comparison are shown in Table~\ref{table:flickr-res}. We reported both image-based and bbox-based mAP; the earlier methods reported only image-based mAP values.

We compared with two existing open-set logo detection approaches~\cite{open-set18,proxy-wacv19}, and three closed-set approaches~\cite{Oliveira2016,Su2016,Bao2016}, all of which are based on the Fast/er R-CNN object detection. Some of these methods trained their models on their own dataset or an expanded version of Flickr-32, since Flickr-32 training set is tiny (only 320 images), making it difficult to train deep networks. Hence, there is some discrepancy in both the training datasets and the networks used among the previous methods. This was mainly due to the lack of a publicly available large-scale benchmark dataset for open-set logo detection; our dataset will fill in this gap.

Without any fine tuning on the Flickr-32 dataset, our SDML with binomial deviance (bindev) loss achieved an image-based mAP of $91.2$, which outperformed the existing best open-set method~\cite{proxy-wacv19} by a large margin, $+34.6$ pps. It even outperformed the best closed-set method by $+7$ pps, clearly demonstrating its effectiveness.

When we fine tune the RFBNet generic logo detection on the small Flickr-32 training set, the bbox-based mAP improved from $52.9$ to $60.3$ ($+7.4$ pps). This indicates that generic logo detection is less transferrable between the two datasets, while logo matching transfers much better. We could not fine tune SDML on Flickr-32 for matching, as it does not have labeled pairs.

We presented end-to-end logo detection samples from OSLD and Flickr-32 test sets in Appendix~\ref{appx:samples}. Our detector is able to detect and recognize transformed logos, even though the canonical logos are upright. The matching network is able to learn how to match transformed logo instances with the help of such samples in the training set. We also experimented with additional data augmentations (random rotations, affine transformations), but did not observe performance improvement.

\noindent
\textbf{Discussion.} The major performance boost in our logo detector comes from the SDML training and matching. The earlier open-set logo detection method in~\cite{open-set18} used Faster R-CNN for detection and the outputs of a classification network for matching, which is not optimal for matching. Proxy-NCA method~\cite{proxy-wacv19} used Faster R-CNN for detection and Proxy-NCA metric learning; trained their networks on a very large dataset of product images (296K images, 2K logo classes), which resulted in $+10$ points lift in the mAP over \cite{open-set18}.

Our method is most similar to \cite{proxy-wacv19}, which also trained on a dataset of product images downloaded from Amazon, but 14 times larger than ours (296K vs 20K) and half the number of logo classes (4.1K vs 2K); their Faster R-CNN mAP and recall are similar to ours, indicating that the major performance boost in our logo detector comes from the SDML training and matching.

As for the closed-set logo detectors~\cite{Oliveira2016,Su2016,Bao2016} which all used Fast/er R-CNN object detector, the major limitation seems to be the Flickr-32's small training set. These methods are not scalable to large number of classes (requiring large amount of memory during training) even when there is sufficient amount of training data for each class, which is very expensive. Moreover, they can not recognize new logo classes not seen during training.

\section{Conclusions}
We presented a two-stage open-set logo detection system (OSLD) that can recognize new logo classes without re-training. We constructed a new open-set logo detection dataset (OSLD) with 12K logo classes, and released it for research purposes. We evaluated OSLD on this dataset and on standard Flickr-32 dataset, demonstrated good generalization to unseen logo classes and outperformed both open-set and closed-set logo detection methods by a large margin.

There is still ample room for improvement in bbox-based metrics, especially with better and more transferrable generic logo detectors, larger and higher quality training sets, and self/semi-supervised learning.

{\small
	\bibliographystyle{ieee}
	\bibliography{references}
}

\clearpage

\appendix

\section{Qualitative Results}
\label{appx:samples}

\begin{figure}[h!]
	\centering	
	\begin{tabular}{c c}
		Ground Truth & Detections \\
		\includegraphics[height=0.43\textheight]{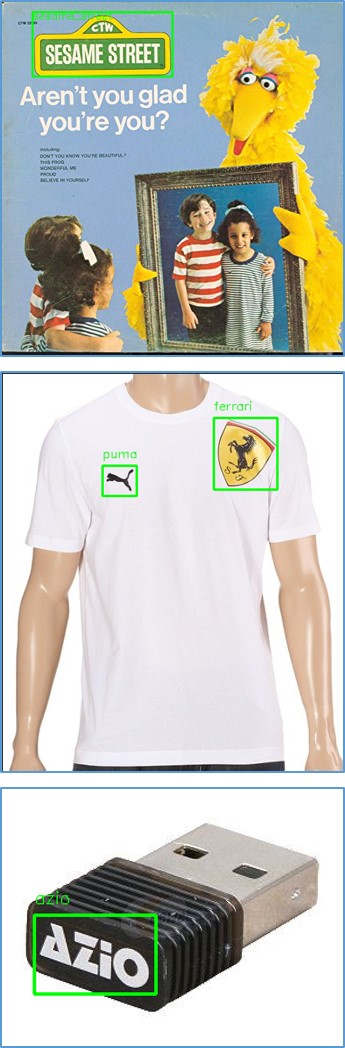} &
		\includegraphics[height=0.43\textheight]{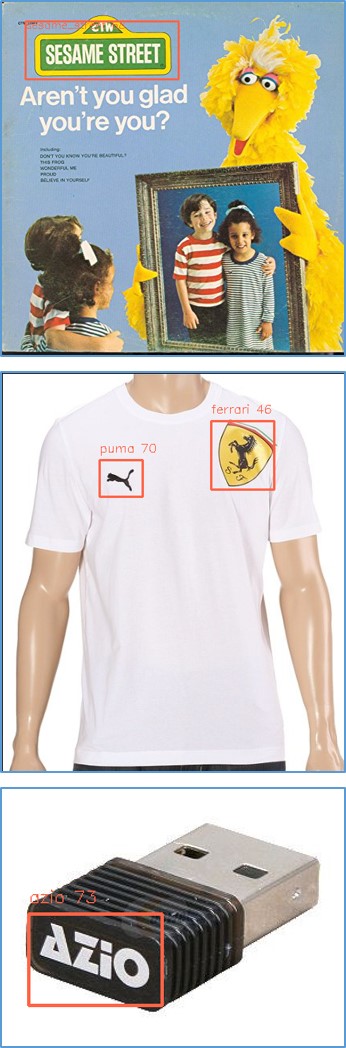} \\
		\includegraphics[height=0.43\textheight]{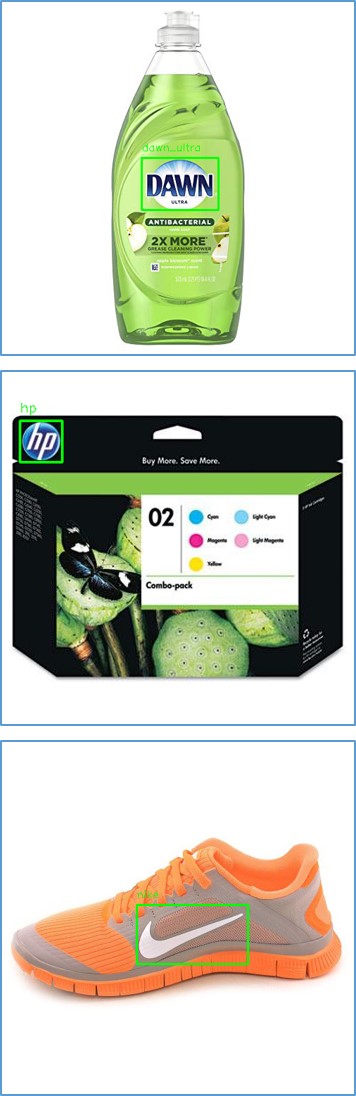} &
		\includegraphics[height=0.43\textheight]{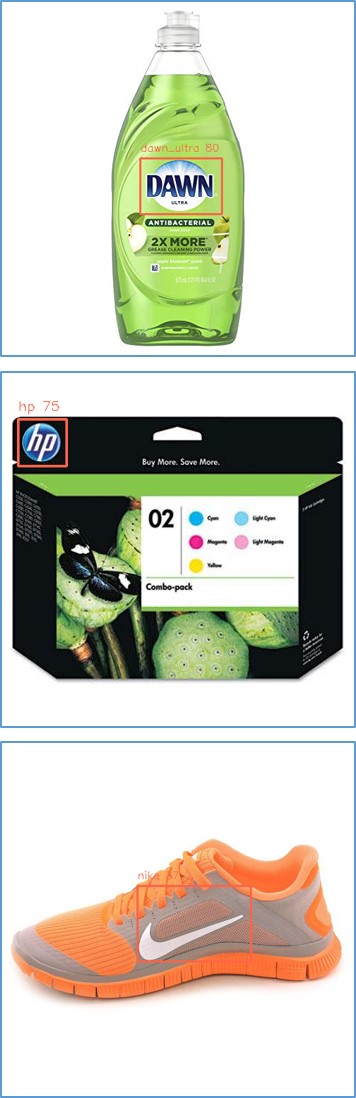}	
		
	\end{tabular}
	
	\caption{End-to-end  logo detection examples on OSLD test set (SDML with binomial deviance loss and 12.1K classes). The numbers show the matching confidence (cosine similarity) out of 100.}
	\label{fig:samples-osld}
\end{figure}

\begin{figure}[h!]
	\centering	
	\begin{tabular}{c c}
		Ground Truth & Detections \\
		\includegraphics[height=0.44\textheight]{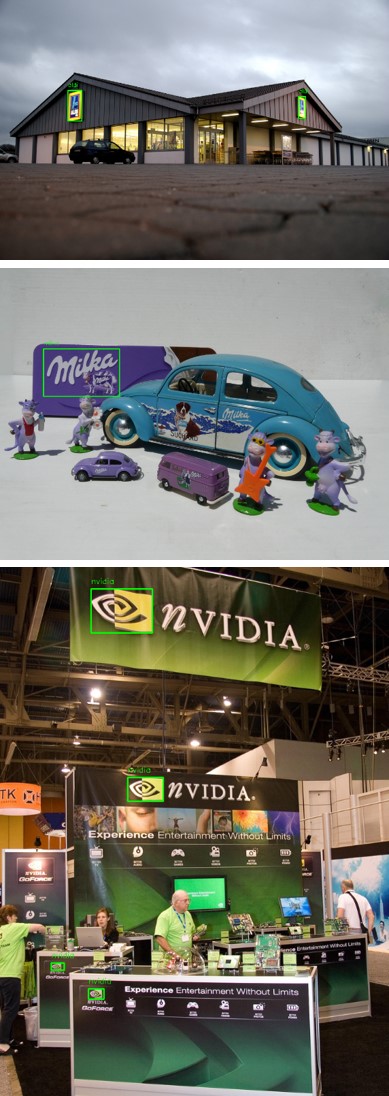} &
		\includegraphics[height=0.44\textheight]{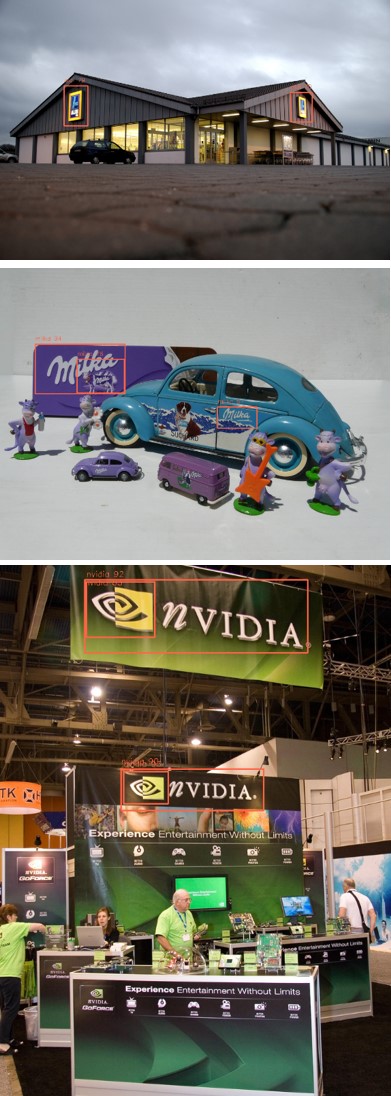} \\
		\includegraphics[height=0.44\textheight]{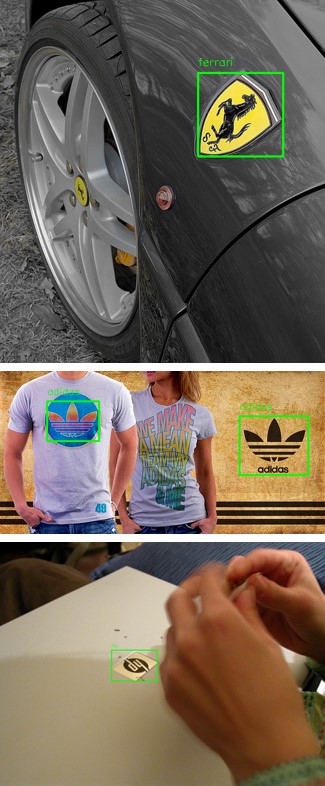} &
		\includegraphics[height=0.44\textheight]{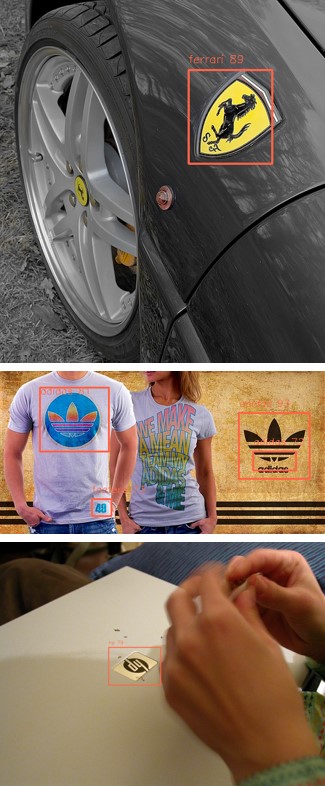}	
		
	\end{tabular}
	
	\caption{End-to-end  logo detection examples on Flickr-32 test set (SDML with binomial deviance loss and 32 classes). The numbers show the matching confidence (cosine similarity) out of 100.}
	\label{fig:samples-fl32}
\end{figure}

\end{document}